\documentclass[12pt]{article}

\usepackage{newtxtext,newtxmath}

\usepackage{graphicx}

\usepackage[letterpaper,margin=1in]{geometry}
\newif\ifIncludeComments
\IncludeCommentsfalse 
\IncludeCommentstrue 
\ifIncludeComments
    \newcommand{\JDC}[1]{{\color[rgb]{1.0,0.0,1.0} (JDC) #1}}
    \newcommand{\YZ}[1]{{\color[rgb]{0.8,0.3,0.3} (Y.Z.) #1}}

    \newcommand{\CS}[1]{{\color[rgb]{0,0.0,1.0} (C.S) #1}}
\else
    \newcommand{\JDC}[1]{\ignorespaces}
    \newcommand{\YZ}[1]{\ignorespaces}

    \newcommand{\CS}[1]{\ignorespaces}
\fi

\renewenvironment{abstract}
	{\quotation}
	{\endquotation}

\date{}

\makeatletter
\renewcommand{\fnum@figure}{\textbf{Figure \thefigure}}
\renewcommand{\fnum@table}{\textbf{Table \thetable}}
\makeatother

\usepackage{scicite}

\usepackage{url}

\newcommand{\coloneqq}{\mathrel{\mathop:}=}
\newcommand{\resistance}{\ensuremath{\alpha_z}}

\def\scititle{
	Scout-Rover cooperation: online terrain strength mapping and traversal risk estimation for planetary-analog explorations
}
\title{\bfseries \boldmath \scititle}

\author{
	S. Liu$^{1\dagger}$, J. D. Caporale$^{1\dagger}$, Y. Zhang$^{1}$, X. Liao$^{1}$, W. Hoganson$^{2}$, W. Hu$^{2}$, \\S. Misra$^{2}$, N. Peddinti$^{2}$, R. Holladay$^{2}$, E. Fulcher$^{1}$, 
    A. Panyam$^{2}$,  A. Puentes$^{2}$, \\J. M. Bretzfelder$^{3,4}$, M. R. Zanetti$^{3}$, 
    U. Wong$^{5}$, D. E. Koditschek$^{2}$, 
 \\M. Yim$^{2}$, D. Jerolmack$^{2}$, C. Sung$^{2\star}$, and F. Qian$^{1\star}$ \and
	\small$^{1}$University of Southern California, Los Angeles, CA, USA.\and
	\small$^{2}$University of Pennsylvania, Philadelphia, PA, USA.\and
    \small$^{3}$NASA Marshall Space Flight Center, Huntsville, AL, USA.\and
    \small$^{4}$Georgia Institute of Technology, Atlanta, GA, USA.\and
    \small$^{5}$NASA Ames Research Center\and
	\small$^\star$Corresponding author. Email: feifeiqi@usc.edu\and
	\small$^\dagger$These authors contributed equally to this work.
}

\begin{document} 

\maketitle

\begin{abstract} \bfseries \boldmath

Robot-aided exploration of planetary surfaces is crucial for advancing our understanding of the geologic processes affecting planetary bodies. However, many high-value scientific sites, such as Martian dunes and lunar craters, present significant challenges due to loose, deformable regolith, making them hazardous for traditional wheeled rovers. Developing innovative strategies to safely explore challenging planetary surfaces can unlock new  exploration opportunities and improve our understanding of planetary environments. 

Our study presents a novel approach that employs a hybrid team of wheeled and legged rovers to expand access to challenging planetary surfaces while mitigating exploration risks. In this approach, high-mobility legged robots serve as ``scouts'' that traverse ahead of wheeled rovers to characterize regolith terramechanical properties, assess rover traversability, and identify scientifically interesting sampling locations to guide safe, efficient sampling operations. 

Here we present our experimental results from two planetary analogue environments --- the NASA Ames  Lunar Simulant Testbed, and the White Sands Dune Field in New Mexico --- to demonstrate the following new capabilities: (1) online terrain regolith mapping using legged scouts; and (2) safe and efficient traverse planning for different rovers. For (1), we demonstrated that the scouting robots were able to use force signals from their legs to build a map of terrain regolith strength using input from each step. For (2), we demonstrated the ability for wheeled and legged rovers to assess traversal risk based on the estimated terrain regolith map, and plan safe exploration paths to reach scientifically-interesting target locations.  

Our results show that this complementary scout-rover cooperation strategy leverages the strengths of different robot platforms to enable robust assessment of regolith mechanical properties, improve identification of scientifically relevant sites, and facilitate safer rover path planning. This hybrid scouting framework has the potential to increase scientific return and mission success in complex planetary environments. 
\end{abstract}


\section{Introduction} \label{Introduction}
Robotic platforms have become central to planetary exploration, enabling the traversal and investigation of remote, extreme environments. To date, most missions have relied on wheeled systems --- such as remotely operated Martian rovers~\cite{farley2020mars, grotzinger2012mars, lindemann2006mars} and both crewed and autonomous lunar vehicles~\cite{carrier1991physical, costes1972mobility, florenskii1978floor, ding20222}. 
Wheeled rovers offer superior energy efficiency on firm ground, mechanical simplicity, structural robustness, and proven flight heritage, making them well suited for sustained operations, complex sampling tasks, and long-duration scientific missions. However, these platforms can face critical limitations across highly deformable and rugged surfaces. Due to the limited degree of freedom in wheels, they have lower ability to actively manage sinkage or recover from entrapment~\cite{florenskii1978floor, ding20222} in loose granular media~\cite{costes1972mobility,david2005opportunity, reid2020mobility} especially when compared to alternative locomotion platforms such as legged or other articulated robots~\cite{vaquero2024eels, 4526511, doi:10.1126/scirobotics.ade9548, valsecchi2023towards}. Without accurate assessment of terrain physical properties and traversal risks, these mobility constraints can lead to mission-threatening failures~\cite{callas2015mars} or overly conservative drive plans that limit access to scientifically-valuable targets~\cite{qiao2021ina,glotch2021scientific, potts2015robotic, steenstra2016analyses, seeni2010robot}, resulting in missed opportunities. Current rover planning relies heavily on remote sensing data --- such as LiDAR, thermal imagery, and ground-penetrating radar~\cite{shepard2001roughness, knight2001ground} --- to infer terrain properties. These methods often yield uncertain and inaccurate estimates of regolith characteristics~\cite{jia2021regolith, aussel2025global},  impeding safe path planning and limiting informed decision-making in navigation and sampling.  

To address this issue, we propose to use high-mobility legged robots as scouts to directly assess regolith properties through locomotion and complement rover operations. 
Legged robots are typically limited in payload capacity as compared to wheeled rovers, but could offer advantages in traversing  deformable and rugged terrains~\cite{li2009sensitive,li2010effect,qian2013walking,johnson_tail_2012,johnson2013leaping,qian2015principles,qian2015dynamics,qian2015anticipatory,qian2017ground,saranli2001rhex,roberts_desert_2014,chenli2018gap,chenli2018bump,qian2019obstacle,lee2020learning, miki2022learning, choi2023learning, Hu2024obstacle,hoeller2024anymal}. In addition, their physical interaction with the terrain during each step provides a unique and versatile sensing modality: every footfall generates direct measurements of ground properties~\cite{qian2019rapid,bush2023robotic,fisher2023lassie,ruck2023downslope,fulcher2024making}. 
Recent advances in actuator technology --- particularly in direct-drive and quasi-direct-drive actuators with minimal gearing backlash --- has enabled high actuator transparency and proprioceptive precision~\cite{kenneally2016design, 10.1007/978-3-030-33950-0_42}. This allows legged robots to accurately measure ground reaction  forces and infer terrain mechanics~\cite{qian2019rapid,bush2023robotic,ruck2023downslope}.
Unlike visual-based terrain assessment~\cite{wu2019tactile, bednarek2019touching, haddeler2022traversability, fahmi2022terrain, ren2024top}, which can struggle to detect deformable or subsurface hazards, interaction-based sensing provides rich information about the mechanical properties of homogeneous and heterogeneous granular terrain, including penetration resistance, shear strength, layering, and cohesion~\cite{qian2019rapid,ruck2023downslope,liu2025adaptive}. 
These directly-sensed terrain parameters are critical for evaluating the traversal risk of different rover configurations and for informing safe and efficient path planning. By pairing legged scouts with wheeled rovers, we leverage the complementary strengths of both platforms: legged robots provide dense, physics-grounded terrain measurements, while wheeled rovers execute payload-intensive scientific operations along routes informed by those measurements. This cooperative paradigm enables terrain-aware operations that expand the reachable science workspace without sacrificing the robustness and efficiency of wheeled platforms.

To demonstrate this potential, we present \textbf{LSRC: Legged Robot Scouting for Rover Cooperation} --- a framework that leverages the unique force-sensing capabilities of legged robots to construct high-resolution regolith strength maps, and converting them into risk maps to guide the exploration planning of heterogeneous rover teams (see Supplementary Video). We test the performance of LSRC through deployment in two planetary analogue environments: the Solar System Exploration Research Virtual Institute (SSERVI) Regolith Lab at the NASA Ames Research Center, and the dunes of White Sands National Park in New Mexico, USA. 

Our first planetary analogue test was deployed at NASA Ames SSERVI Regolith Lab, which is a controlled lunar-analogue environment  with high-fidelity lunar simulant LHS-1~\cite{yin2023shear}, designed to support rover testing in conditions representative of the Moon’s polar regions. The testbed allows systematic control of regolith properties and serves as an ideal setting to assess the accuracy of terrain sensing through locomotion. We prepared the terrain at three compaction levels and quantified regolith strength using two methods: (1) a stationary, robotic-leg penetrometer (serving as ground truth) and (2) the legged scout robot, Spirit (Sec. \ref{sec:Ames-setup}). We find that the regolith strength maps generated by the legged scout closely reflect the ground truth, demonstrating the feasibility of using locomotion as a sensing modality in planetary analogue environments (Sec. \ref{sec:ames-terrain-mapping}). Building on this regolith strength map, we further demonstrate in laboratory experiments that these estimated terrain properties are directly linked to the traversability of wheeled and RHex-class~\cite{saranli2001rhex} rovers, including observed metrics such as wheel slippage and leg sinkage (Sec. \ref{sec:ames-risk-estimation}). By integrating regolith strength with locomotion models~\cite{li2009sensitive, ding2020definition}, we generate traversal risk maps that can inform safe and efficient exploration planning. 

Building on the validated terrain-mapping capabilities from our first analogue deployment, we conducted a second planetary analogue mission at White Sands National Park, New Mexico  (Sec. \ref{sec:WS-setup}) --- an active gypsum dune field that closely mimics the aeolian terrains encountered at Meridiani Planum by the NASA Opportunity rover~\cite{grotzinger2005stratigraphy, chavdarian2006cracks, jerolmack2006spatial}. In addition to loose sand and salt crusts, representative of Opportunity observations, there are also abundant microbial mats and crusts, which contribute to significant rover mobility challenges and offer distinct scientific targets. This combination of features makes White Sands an ideal analogue for evaluating our system's task planning and safe navigation performance based on scouted terrain maps. During this field campaign, the legged robots performed regolith strength mapping for a 10 m by 15 m 2D transect, and constructed a traversal risk map to guide wheeled rover traversal (Sec. \ref{sec:WS-mapping}). Our mission planner then successfully directed the rover to the scientific targets while avoiding high-risk areas (Sec. \ref{sec:WS-navigation}). In contrast, when operating without access to the scouted terrain information, the same wheeled rover failed to reach the scientific targets due to sand entrapment. Our results demonstrated the effectiveness of our cooperative scouting framework in guiding rovers through challenging planetary-analogue terrains using the scouted risk map, paving the way for future deployment in extraterrestrial environments.



%

\section{Results}
\subsection{Analogue Mission Overview: Scouting-Rover cooperation for planetary-analogue exploration}\label{sec:results-overview}

\begin{figure*}[hb!]
    \centering  
    \includegraphics[width=0.9\linewidth]{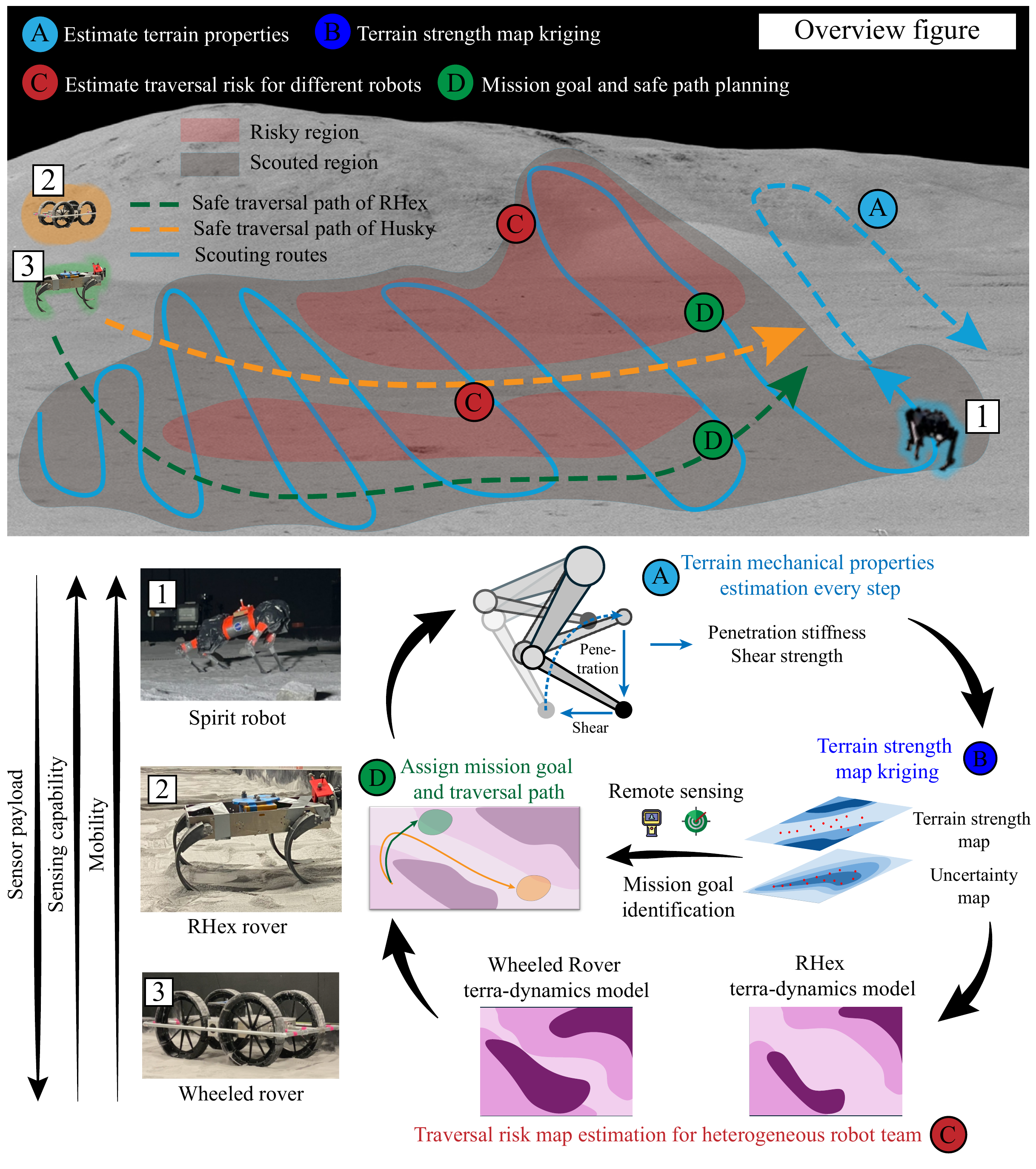}  
    \vspace{-0.4in}
    \caption{Heterogeneous robot team framework for terrain-aware planetary exploration.
(A) A legged scout robot estimates terrain properties using proprioception every step. (B) The terrain properties are used to generate terrain strength and uncertainty maps via kriging. (C) Robot-specific terradynamics models are used to estimate traversal risk for RHex rover and wheeled rover with different morphology and sensing payloads. (D) A central planner assigns mission goals and traversal path based on scientific targets and terrain strength maps, ensuring safe operation. }
    \label{fig:overview}  
\end{figure*}

Our mission framework employs a heterogeneous team of ground robots to enable robust, terrain-aware, and science-driven planetary exploration. The team consists of three mobile platforms: a highly-mobile legged scout robot, Spirit (Fig. \ref{fig:overview}, robot 1), a RHex-style legged rover (Fig. \ref{fig:overview}, robot 2), and a wheeled rover (Fig. \ref{fig:overview}, robot 3). The hybrid legged and wheeled rovers provide complementary advantages in mobility (traverse many types of terrain), strength (can carry sensor payloads), and sensing (estimating risk from ground interactions). 

When cooperatively exploring an unknown region, the legged robot is deployed first to scout the area (Fig. \ref{fig:overview}, blue line), leveraging its proprioceptive abilities to assess terrain properties during locomotion. As it walks, the legged scout collects discrete measurements of regolith strength (Fig. \ref{fig:overview}A), which are fused to generate a continuous, spatially-resolved terrain map via Gaussian Process Regression (GPR) \cite{seeger2004gaussian, vasudevan2009gaussian} (Fig. \ref{fig:overview}B). We integrate the terrain map with resistive force theory~\cite{li2013terradynamics} and platform-specific locomotion models~\cite{qian2015principles, agarwal2021surprising, shrivastava2020material} to estimate the traversal risk for each rover type (Fig. \ref{fig:overview}C). In parallel, the same terrain measurements properties can be used to infer regions of scientific interest by evaluating discrepancies between sensed and predicted regolith properties under a prior hypothesis~\cite{liu2023understanding}. This process yields a “science reward” map (Fig.~\ref{fig:overview}D), which can facilitate the selection of scientific targets. The identified targets and traversal risk map can then be considered jointly to inform safe rover path planning while supporting scientific objectives.

To navigate toward these scientific targets while avoiding hazardous terrain, we employ a potential field-based planner~\cite{arslan2019sensor, vasilopoulos2022reactive, rousseas2024reactive}. The planner balances between reward (represented as attractors) and risk (represented as a repulsive field) to select safe, high-value paths for each platform. While the wheeled and RHex rovers execute their tasks along these paths, the scout robot can continue to expand the terrain map opportunistically, updating the risk map in real time.
This cooperative exploration strategy enables informed, safe navigation while maximizing scientific return --- a key capability for future surface missions targeting diverse and uncertain environments.


\subsection{Analogue Mission 1 at the NASA Ames Lunar Simulant Testbed: Regolith Strength Mapping and Traversal risk Estimation}\label{sec:Ames-results}

\subsubsection{Regolith Preparation and Strength Ground Truthing}\label{sec:Ames-setup}

\begin{figure*}[hb!]
    \centering  
    \includegraphics[width=1\linewidth]{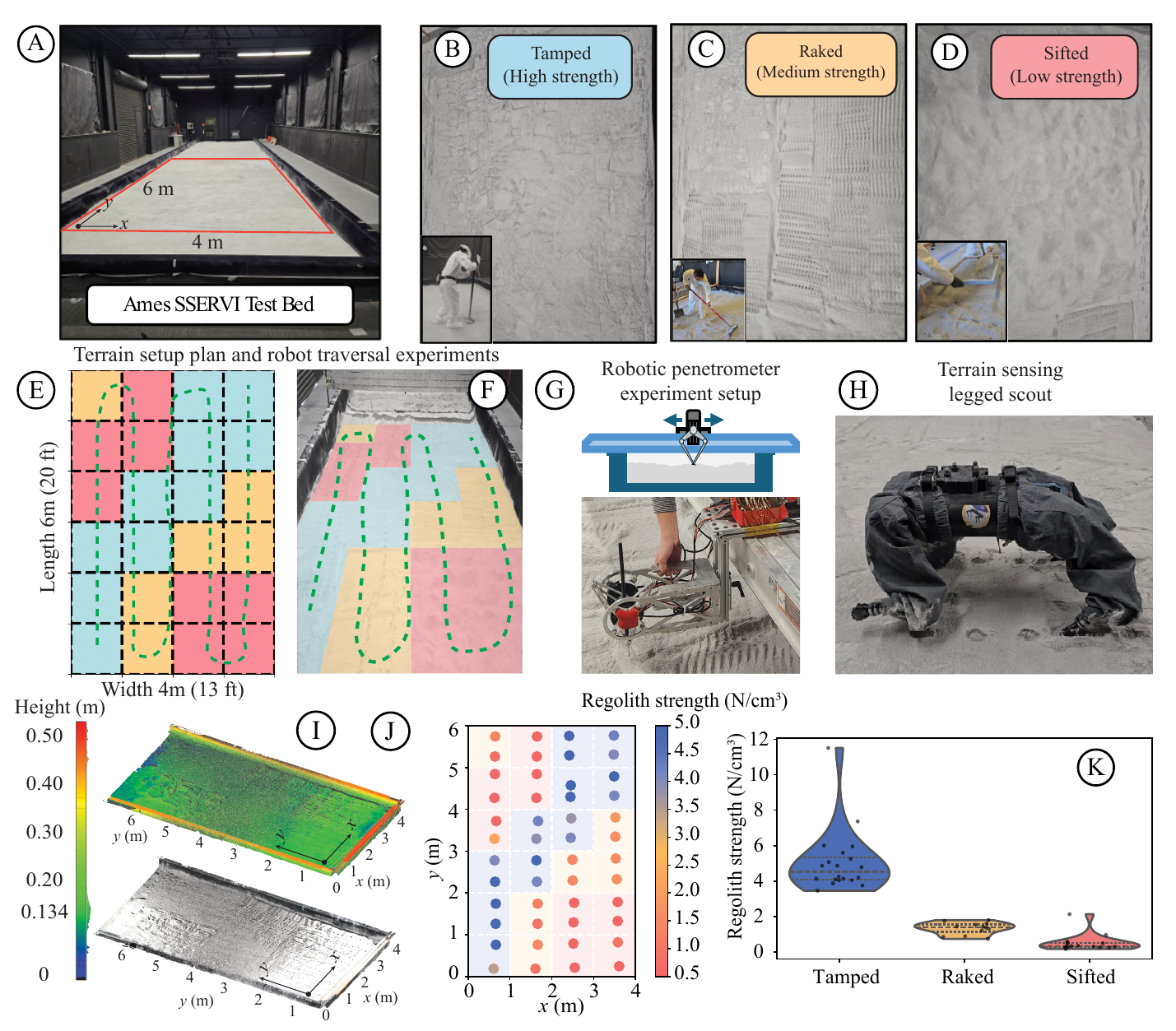}  
    \caption{Experimental setup for terrain-aware legged robot scouting.
(A) Experiment site, the SSERVI Regolith Lab testbed located at the NASA Ames Research Center. (B–D) Three terrain-preparation protocols with strength from high to low: tamped (high strength), raked (medium strength), and sifted (low strength). (E)(F) The desired terrain strength map, prepared using the three protocols. (G) Ground truth terrain strength measurements are collected using a robotic penetrometer across the testbed. (H) A legged scout robot performs terrain scouting using proprioception from every step along its path. (I) LiDAR scan after terrain preparation. (J) Penetration resistance measurements from the robotic penetrometer across the testbed. (K) The penetration resistance distribution from terrain grid cells prepared using the three protocols.} 
    \label{fig:setup}  
\end{figure*}

We deployed our regolith mapping and traversal risk estimation module in a 6-meter by 4-meter section of the Ames SSERVI Regolith Lab. The testbed was filled with LHS-1 lunar simulant to an average depth of 7--8\,cm (Fig.~\ref{fig:setup}A) to minimize bottom boundary effects that can influence force transmission in shallow granular beds~\cite{stone_getting_2004}. To generate controllable variation in regolith strength and resulting variation in traversal risk, the simulant was divided into a 1-meter grid (Fig.~\ref{fig:setup}E, F). Each cell was manually prepared to one of three compaction levels by applying distinct surface treatments: \textbf{(1) tamping}, which prepares regolith to high compaction and strength by firmly compacting the simulant using a hand tamper (Fig.~\ref{fig:setup}B); 
\textbf{(2) raking}, which prepares regolith to medium compaction and strength by loosening and evenly redistributing the simulant surface using a garden rake (Fig.~\ref{fig:setup}C); and 
\textbf{(3) sifting}, which prepares the simulant to low compaction and strength by passing them through a mesh sieve (Fig.~\ref{fig:setup}D). Using these three treatments, the testbed was prepared to a pre-designed spatial variation pattern of regolith strength 
as shown in Fig.~\ref{fig:setup}E, F.
After the regolith was prepared following the predesigned pattern, the surface appeared relatively flat, with height standard deviations of 1.66 centimeters according to the LIDAR scan (Fig.~\ref{fig:setup}I). Visually, it was difficult to discern the differences in terrain traversability.  
To determine the strength across the testbed, we collected ground-truth regolith strength measurements using a robotic leg penetrometer, Traveler (Fig.~\ref{fig:setup}G). The Traveler toe penetrates vertically into the regolith, while measuring how quickly the ground reaction forces increase with the penetration depth  (see Sec. \ref{sec:method-penetration resistance}). The slope of ground reaction force with respect to depth -- referred to as penetration resistance -- is a commonly-used metric for assessing regolith normal strength~\cite{li2013terradynamics}. 
The accuracy of the robotic leg penetrometer has been extensively validated in previous studies, and deployed in a wide range of laboratory and field missions to characterize regolith properties~\cite{qian2019rapid,ruck2023downslope,bush2023robotic,liu2025adaptive}. 
Leg penetrometer measurements shown that the three treatments yielded well-separated resistance values, where the penetration resistance per unit area of the tamped, raked, and sifted terrains was measured as 5.07~$\pm$~1.79 N/cm$^3$, 1.34~$\pm$~0.35 N/cm$^3$, and 0.51~$\pm$~0.47 N/cm$^3$, respectively (Fig. \ref{fig:setup}K). Despite small variability due to manual terrain preparation, the measured resistance values clearly distinguished the three terrain conditions and confirmed the intended distribution pattern (Fig.~\ref{fig:setup}J).


\subsubsection{Regolith Strength Mapping Results}\label{sec:ames-terrain-mapping}

After ground-truth measurements of the prepared regolith were taken, a legged scout (Fig.~\ref{fig:setup}H) was deployed to construct the regolith mechanics map. The regolith mechanics map captures the strength of unconsolidated regolith in response to intrusion forces from robot legs and rover wheels, and the spatial uncertainty in strength estimation. These regolith mechanics measurements directly reflect key substrate properties, such as packing density~\cite{qian2015principles}, moisture content~\cite{qian2019rapid}, grain size~\cite{ruck2024downslope} and layering~\cite{bush2023robotic}, and therefore are essential for estimating traversal risk and guiding safe navigation. 

\begin{figure*}[hb!]
    \centering  
    \includegraphics[width=0.95\linewidth]{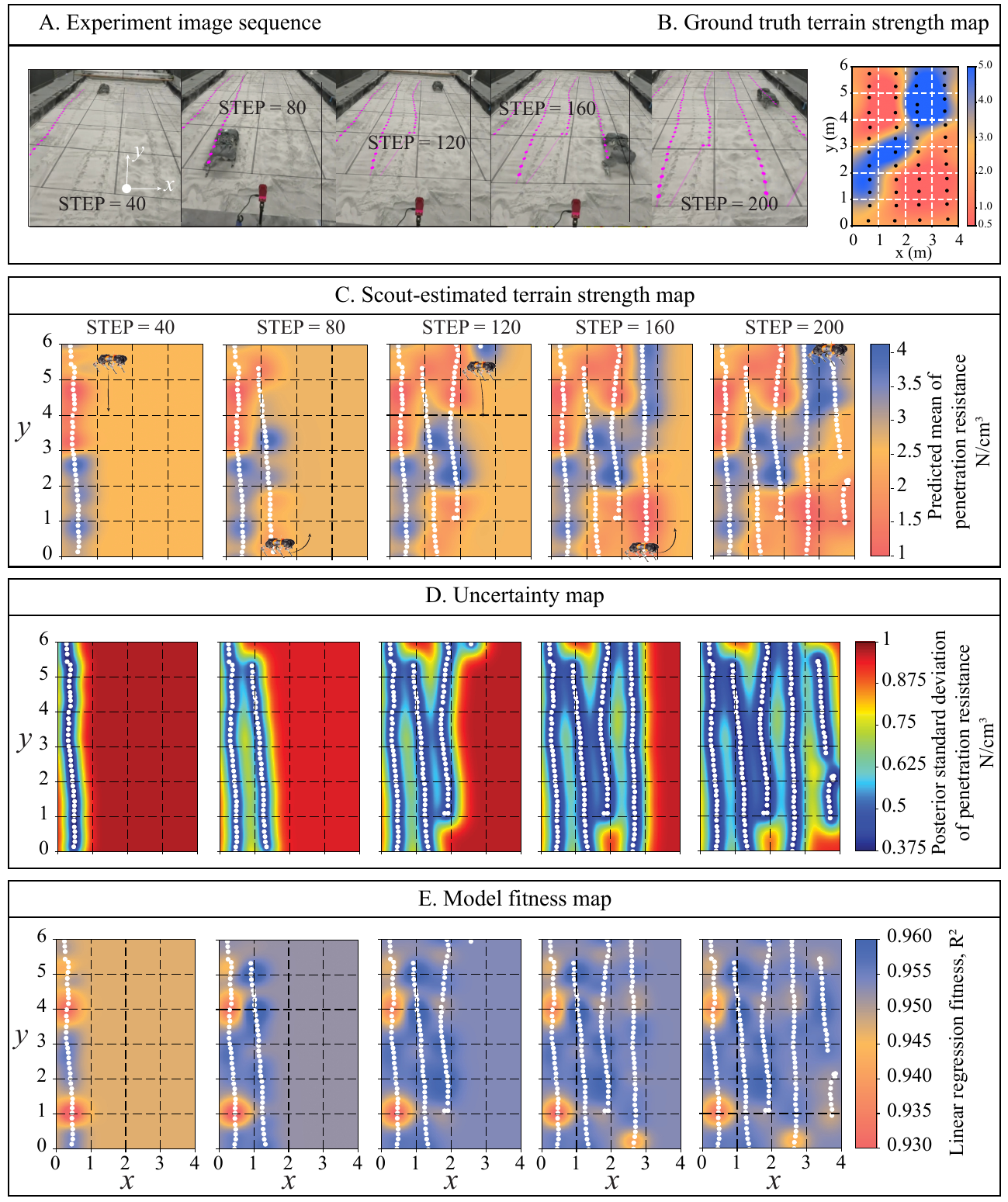}  
    \caption{Online terrain property mapping during robot traversal.
(A) Robot snapshots at successive time steps. (B) Penetration resistance ground truth measured by Traveler. (C) Sequential terrain property maps inferred online by the scout robot, Spirit, from proprioceptive sensing; white markers denote the robot trajectory and inset icons indicate movement direction. (D) Corresponding Gaussian process–based uncertainty maps. (E) Coefficient of determination of the linear terrain mechanics model.} 
    \label{fig:terrainmap}  
\end{figure*}

To produce the regolith mechanics map, the scout used a customized Crawl-N-Sense gait (see Sec. \ref{sec:method-sensing gait}), enabling proprioceptive measurement of ground reaction forces during each step (Fig.~\ref{fig:terrainmap}A). As the scout traversed the testbed along a predefined zig-zag path (Fig.~\ref{fig:setup}E, F, green dashed line), a linear regression was applied to compute penetration resistance (see Sec. \ref{sec:method-sensing gait}), a metric of the regolith strength that directly related to packing density. The coefficient of determination ($R^2$) was used to quantify the degree to which the linear regressions captures the granular rheological behavior. For homogeneous granular media under low-speed intrusion, the linear regression model is expected to capture granular response well~\cite{li2013terradynamics}. 
The estimated penetration resistances, combined with robot toe contact positions computed from the motion-capture system, were used to construct a spatially continuous regolith strength map using Gaussian process regression (Fig.~\ref{fig:terrainmap}C).

We present the regolith strength map and validate it against ground-truth measurements. Across more than 200 steps, the system reliably captured spatial variations in substrate strength, such as the central soft-sifted area (blue shading in Fig. \ref{fig:terrainmap}B, C). The average penetration resistance per unit area are 0.60 $\pm$ 0.25, 1.21 $\pm$ 0.23, and 1.80 $\pm$ 0.24 N/cm$^3$ for the sifted, raked, and tamped regions, respectively. The Crawl-N-Sense measurements closely match the ground truth in the sifted and raked regions. In the tamped region, however, the terrain strength was underestimated, likely due to the robot's rubber toe deformation on relatively rigid substrates, leading to overestimating in penetration depth and therefore underestimation in penetration resistance. 
Despite this limitation, the underestimation in high-strength regions does not compromise risk estimation, since rover failures predominantly occur in softer regions.

Along with the strength map, Gaussian process regression simultaneously produces a spatial uncertainty map (Fig. \ref{fig:terrainmap}D). This uncertainty field provides critical input for safe-path planning algorithms for planetary rovers, enabling risk-aware navigation decisions based on the reliability of terrain predictions. In addition, the scout produces a model fitness map (Fig. \ref{fig:terrainmap}E) using the model fitness value, $R^2$, from each measurement location, to quantify the degree to which each measurement conforms to the homogeneous granular behavior. Regions of low model fitness can be particularly valuable for scientific exploration, as they can signal  unexpected terrain features beyond prior assumption,  such as the presence of surface crusts, subsurface ice deposits, and cohesive layers. Therefore, the model fitness map can be used to identify potential areas of scientific interest~\cite{liu2023understanding}.

These results demonstrated that proprioceptive sensing can offer accurate estimation of terrain penetration resistance, highlighting the legged robot's potential as an effective scout in detecting low-strength regions in deformable granular substrates, where vision-based methods may not be sufficient.

\subsubsection{Traversal Risk Estimation Results and Simulation Validation}\label{sec:ames-risk-estimation}

Building on the terrain strength maps generated by the legged scout, we demonstrate how these measurements enable risk assessment for different rover configurations and payload capacities. Our risk estimation framework integrates terradynamics models with the regolith strength data to predict traversal performance across varying terrain conditions and robot configurations (Sec. \ref{sec:method-risk}).

\begin{figure*}[hb!]
    \centering  
    \includegraphics[width=1\linewidth]{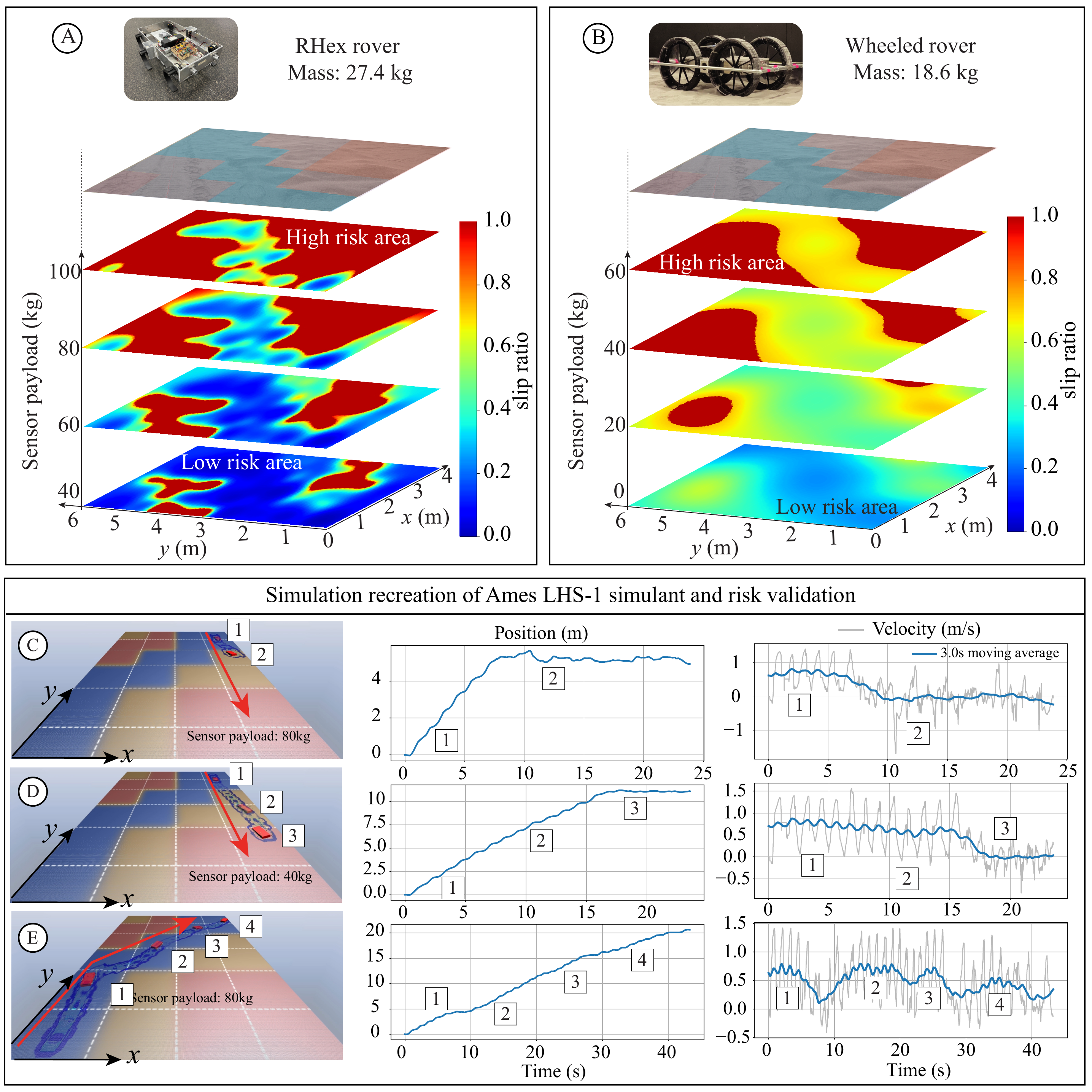}  
    \caption{Traversal risk estimation and simulation validation for planetary rovers. (A) Distribution of RHex slip ratio  within the mapped region, illustrating increasing traversal risk (red areas) with larger sensor payloads (from 40 to 100 kg).  (B) Distribution of slip ratio of the wheeled rover for payload range between 0 kg to 60 kg. (C-E)  Simulation validation in a digital twin of the Ames LHS-1 testbed using Chrono. (C, D) RHex (80 kg and 40 kg) moves along a straight path with decreasing strength. The 80 kg robot becomes immobilized in medium-strength terrain (yellow), whereas the 40kg robot successfully traverses the region without failure. (E) RHex (80 kg) moving along a safe path based on the terrain strength map. For all scenarios, the position (middle) and velocity (right) are recorded over time.} 
    \label{fig:risk}  
\end{figure*}

To estimate traversal risk for each rover type, we employ validated mechanical models that compute performance metrics as a function of terrain regolith strength and robot configuration parameters (Fig. \ref{fig:risk}). For the RHex rover (Fig. \ref{fig:risk}A), our model predicts slip percentage based on leg angular velocity, robot mass, and terrain penetration resistance (see Sec. \ref{sec:risk-RHex}). The risk maps reveal a striking sensitivity to payload variations: increasing sensor payloads from 40 kg to 100 kg significantly affects the traversable landscape (Fig. \ref{fig:risk}A). At low payloads (40 kg), the rover maintains a low slip ratio ($<$ 10\%) across most of the testbed, but as the payload increases, high-slip regions (Fig. \ref{fig:risk}A, red regions) expand substantially, creating substantial mobility constraints that could lead to mission-threatening immobilization ($>$90\%).

For the wheeled rover (Fig. \ref{fig:risk}B), we use Granular Resistive Force Theory (RFT) to predict the slip ratio and the required drive torque, considering factors such as wheel geometry, vehicle mass, and terrain properties (see Sec. \ref{sec:risk-wheeled}). The resulting slip ratio and torque maps highlight a fundamental trade-off: while wheeled rovers can carry heavier payloads, they can get stuck in the immobilization area with a large slip ratio more easily (63\% with 60 kg payload) compared to legged platforms (28\%) due to their greater mass and fundamentally different interaction dynamics with deformable terrain. 

To validate our risk prediction framework, we conducted comprehensive simulation tests using the Project Chrono open-source physics engine~\cite{projectChronoWebsite}, which has been systematically validated against real-world experiments~\cite{Chrono2016}. We created a virtual replica of the NASA Ames LHS-1 testbed by using the ground-truth regolith strength measurements collected by the Traveler leg penetrometer to parameterize the simulation environment. Specifically, we computed the average penetration resistance value for each 1 m $\times$ 1 m cell and mapped these values to corresponding elastic and damping parameters for the SCM (Soil Contact Model) terrain representation used in the simulation~\cite{krenn2009scm} (see Appendix \ref{sec:appendix} for the complete stiffness parameter matrix and mapping methodology). This spatial parameterization creates an 18 m $\times$ 12 m grid where each cell corresponds to a 3 m $\times$ 3 m region of the physical testbed, with penetration resistance per unit area ranging from 0.28 to 8.38 N/cm³ and reflecting the three distinct terrain types (tamped, raked, and sifted), represented by different colors in the simulation visualization (Fig. \ref{fig:risk}, bottom section). Using this simulation environment, we tested three critical scenarios that demonstrate the importance of payload-aware risk assessment.


First, we simulated a RHex rover (27.4 kg) carrying an 80 kg payload navigating across high-, medium-, and low-strength regions (Fig.~\ref{fig:risk}C). As predicted by our model (Fig. \ref{fig:risk} A), the RHex rover experienced nearly 100\% slip in the medium-strength region, causing its velocity to fall to zero and leading to immobilization shortly after entering that zone (Fig.~\ref{fig:risk}C). Next, to test whether reducing payload mitigates risk, we lowered the mass to 40 kg based on our risk map predictions. In this configuration, the rover successfully traversed the medium-strength region before failing in the low-strength region (Fig.~\ref{fig:risk}D). The velocity and position profiles confirm this failure mechanism, showing rapid deceleration and eventual immobilization as the rover entered the weaker terrain. Finally, we explored safety-aware path planning as an alternative risk reduction strategy. With the high payload mass (80 kg), we planned a two-segment route confined to high-strength (blue) terrain regions (Fig.~\ref{fig:risk}E). The rover successfully followed this path with acceptable average velocity (Fig.~\ref{fig:risk}E), confirming the benefits of integrating terrain information into navigation planning.

Together, these results demonstrate that our terrain-property-based model accurately predicts rover performance across different payloads and terrain conditions. Quantifying traversal risk enables mission planners to balance sensor payloads, select appropriate rover platforms, and design safer routes. This capability underscores the critical role of terrain characterization in safety-aware planning for planetary exploration missions.

\subsection{Analogue Mission 2 at White Sands National Park: Safe and Efficient Navigation to Areas of Scientific Interest}

\subsubsection{Site selection and field workflow}
\label{sec:WS-setup}
To demonstrate the advantages of using a scout robot to support scientific target selection and safe rover traversal, we deployed our scout–rover cooperation system at White Sands National Park. White Sands is an active gypsum dune field within a playa–lake basin that displays a rich variety of sedimentary features and surface textures, including loose sand, partially lithified sandstones, and salt and microbial crusts. Features at this site closely resemble the terrains encountered at Meridiani Planum by the NASA Opportunity rover \cite{grotzinger2005stratigraphy, chavdarian2006cracks, jerolmack2006spatial}, and the existing robust body of work characterizing this dune field facilitates our investigations~\cite{pelletier2014multiscale, qian2019rapid, cooke2025evolution}. 

For our experiments, we selected a representative barchan dune area (Fig. \ref{fig:scientifc}A) composed of loosely packed sand, where variations in grain size and cohesion pose significant mobility challenges for wheeled rovers. In addition, the interdune region exhibits a wide range of surface crusts \cite{jerolmack2012internal}, offering distinct scientific targets. The combination of these features makes this site an ideal analogue for evaluating our system’s task planning and safe navigation performance based on the scouted map.

The analogue mission began with satellite imagery analysis (Fig. \ref{fig:scientifc}A), which was used to identify key operational areas and to generate an overall plan with candidate regions to investigate. We additionally collected lidar scans~\cite{Zanetti2023MobileLiDAR, Anzalone2025ArtemisNavigation} of the region (Fig. \ref{fig:scientifc}B, C) to create high-fidelity digital maps of the specific dune morphology (Fig. \ref{fig:scientifc}B, Dune crest, slip face, and interdune region) and identify feasible corridors for rover traversal (Fig. \ref{fig:scientifc}C, regions with slope angle $<$ 15$^\circ$).  

Once the feasible corridors were identified, we deployed the scout robot first to map the regolith strength throughout the potential rover traversal area (Fig. \ref{fig:scientifc}D, E), and determine the distribution of risk level. 
After scouting, the robot reconstructed the traversal risk map based on the wheeled rover parameters (Fig. \ref{fig:scientifc}F). The mission planner then used the risk map to plan a safe traversal path to reach scientific targets while avoiding high-risk areas. 

In the following sections, we present the terrain strength maps and traversal risk estimation (Sec. \ref{sec:WS-mapping}), and rover terrain-aware navigation performance guided by these maps (Sec. \ref{sec:WS-navigation}). This field mission demonstrates the robustness of our approach in guiding rovers through challenging terrain using the scouted risk map, paving the way for deployment in future planetary environments.

\subsubsection{Legged scouting measurements to identify scientific interest and safe regions}\label{sec:WS-mapping}

\begin{figure*}[hb!]
    \centering  
    \includegraphics[width=1\linewidth]{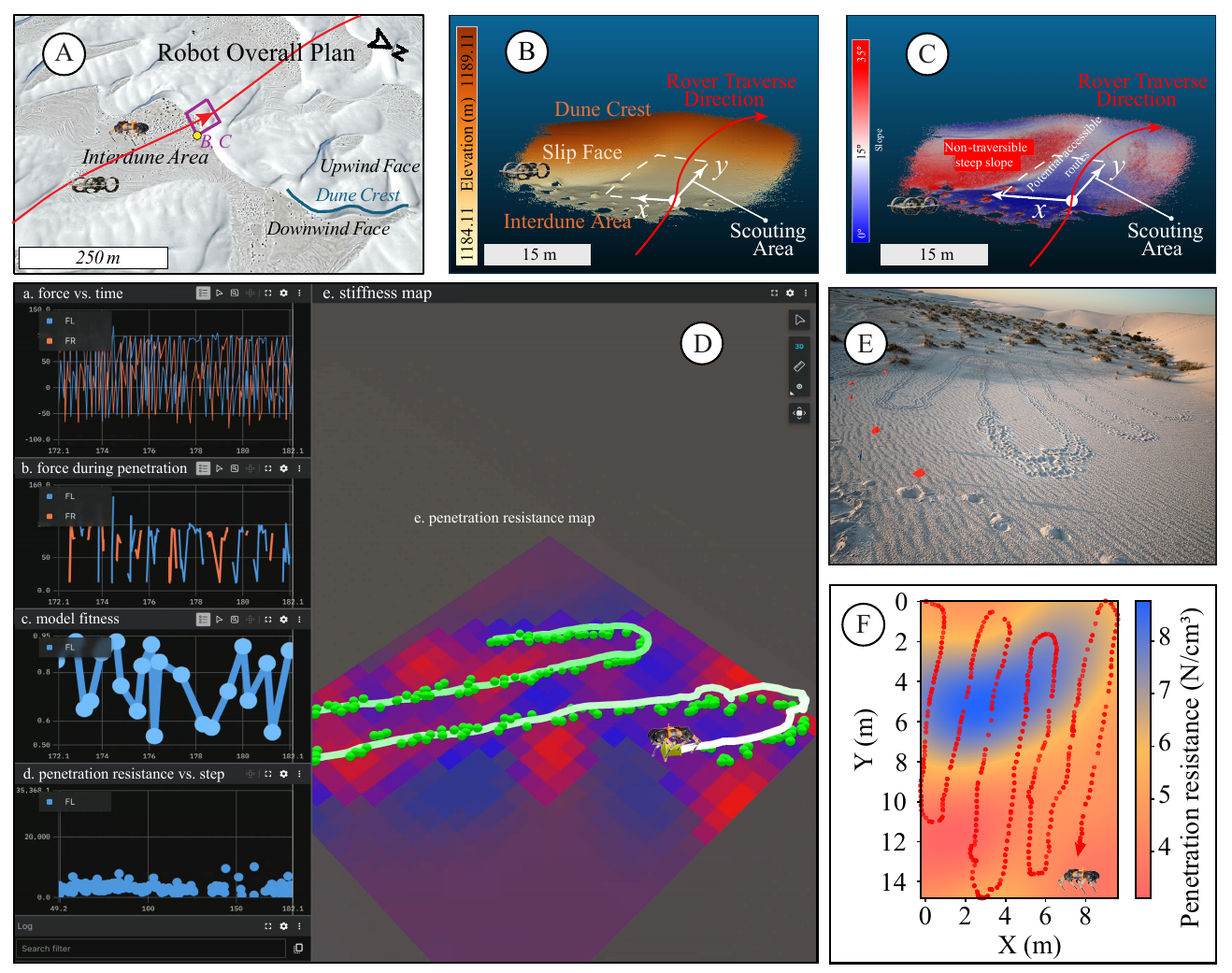}  
    \caption{Workflow for safe rover mission planning integrating satellite imagery, in-situ measurements, and science-interest-driven goal selection.
(A) Overview of the mission plan showing a top-down view of the operational area and the tentative traversal region (purple rectangle).
(B) LiDAR-based reconstruction of the traversal region with elevation data (1184–1189 m).
(C) LiDAR-based inclination measurements, and initial identification of candidate rover routes.
(D) Field interface enabling real-time visualization of terrain strength and science goal selection: (a) estimated toe force over time; (b) toe force during the penetration phase; (c) model fitness; (d) penetration resistance measured from each step; and (e) interpolated penetration resistance map generated via Gaussian Process regression.
(E) Field snapshot after Spirit robot scouting.
(F) Final penetration resistance map of the scouted region.} 
    \label{fig:scientifc}  
\end{figure*}

The scout robot followed a boustrophedon (zig-zag) trajectory (Fig. \ref{fig:scientifc}E, and Fig. \ref{fig:scientifc}F red dashed lines) to characterize the regolith strength. At each step, the scout logged GPS coordinates (Fig. \ref{fig:scientifc}D-e, green markers), along with proprioceptive measurements of penetration force and depth from its front legs (Fig. \ref{fig:scientifc}D-a, b). These measurements were processed online to produce terrain penetration resistance values (Fig. \ref{fig:scientifc}D-d), generating a real-time terrain strength map (Fig. \ref{fig:scientifc}D-e, blue to red colors). In parallel, a model fitness map quantified by $R^2$ (Fig. \ref{fig:scientifc}D-c) was generated to flag potentially interesting regions, such as surface crusts~\cite{bush2023robotic}, subsurface ice~\cite{ruck2024unveiling}, or cohesiveness~\cite{ruck2024downslope}, where force responses deviated from predictions of homogeneous granular models. The flagged regions were then reviewed by planetary scientists with expertise in the geologic processes relevant to the site, and we leveraged their scientific interpretations and analyses to select targets.  

Integrating risk maps with identified scientific targets provided a comprehensive basis for mission planning, enabling rovers to reach high-priority sites while avoiding mobility hazards. Building on the complete terrain strength maps generated by the legged scout (Fig. \ref{fig:scientifc}F), we convert these measurements to rover traversal risk levels (Fig.~\ref{fig:scientifc}G) to guide safe navigation. The risk evaluation considered the predicted slip ratio and the required drive torque of the wheeled rover, estimated based on terrain strength, rover wheel size, and payload capacity (see Sec. \ref{sec:risk-wheeled}). These metrics were used to estimate likelihood of immobilization, providing a quantitative measure to ensure safe navigation across varying terrain conditions. 

In the following section, we report the risk-aware rover navigation performance based on the traversal risk information.

\subsubsection{Terrain-aware rover navigation based on scouting measurements}\label{sec:WS-navigation}
The integration of risk maps and selected scientific interest goals provides rich information for mission planning that allows rovers to safely navigate to scientific targets while avoiding high-risk areas. To demonstrate this capability, we used a reactive potential field planner~\cite{vasilopoulos2022reactive, arslan2019sensor}, 
in which goal locations corresponding to  high scientific reward generate attractive forces that pull rovers closer and the boundaries of high-risk regions generate repulsive forces that push rovers away from hazardous terrain. 
The path that the rover takes is then dictated by the sum of attractive and repulsive forces acting on the rover at each point in time.
This approach ensures that the rover navigates along a path that achieves scientific discovery goals while minimizing the probability of immobilization.
Note that for the purposes of path planning, the rover was modeled as a point mass with a bounded acceleration.

To validate our approach, we conducted field experiments comparing a naive path with the safe path generated by our potential field planner (Fig. \ref{fig:planning}). The naive path ignores terrain risk and moves the rover straight to each goal sequentially, while the safe path strategically avoids high-risk regions. The rover was  teleoperated to follow each path sequentially. For the naive path (Fig. \ref{fig:planning}A),  the rover successfully followed a straight path to the first goal but was unable to reach the second goal. The location at which it started to veer off-course (Fig.~\ref{fig:planning}C.\textcircled{2}) and eventually stopped (Fig.~\ref{fig:planning}C.\textcircled{4}) correspond to regions of increasing risk  that the rover would not be able to move through without higher torque wheel inputs. In contrast, for the safe path (Fig. \ref{fig:planning}B),
the rover was able to  successfully navigate the terrain by strategically avoiding the predicted high-risk region. The rover maintained relatively constant speed throughout the path, except in the mid-risk region (Fig.~\ref{fig:planning}C.\fbox{3}), where it needed to slow down in order to make the sharp turn necessary to avoid the high risk region. The results validate the effectiveness of our risk-aware planning approach in real-world challenging terrain.

\section{Discussion}

This work demonstrates that legged robots can function not only as mobility platforms, but as active scientific scouts that directly sense and characterize terrain mechanics through locomotion. By converting proprioceptively measured leg–terrain interaction forces into regolith properties and traversal risk maps, our framework enables a fundamentally different mode of rover operation: one that is informed by dense, direct surface measurements rather than indirect remote sensing alone. Across both controlled laboratory conditions and a large-scale planetary analogue field deployment, our results show that scouted regolith strength maps reliably reflect spatial variability and meaningfully predict rover traversal risks. These findings highlight the value of embodied interaction as a sensing modality and establish legged–wheeled cooperation as a practical strategy for improving safety, efficiency, and scientific reach in planetary exploration. 


Beyond mobility risk assessment, this scout–rover cooperation framework opens new opportunities for science-driven exploration planning. Measured regolith properties can be compared against scientific hypotheses and remote sensing predictions to identify regions with large hypothesis–measurement discrepancies that warrant further investigation~\cite{liu2024multiobjective, liu2023understanding}. Jointly considering scientific value and traversal risk enables rovers to optimize both their paths and stopping locations, moving beyond conservative, hazard-avoidance strategies toward deliberate, information-rich exploration of mechanically complex terrains.



The traversal risk maps produced by the parameterized terrain model further enable adaptive mission-level decision-making. For example, payloads could be distributed across multiple rovers based on the measurements required at each scientific target and the associated payload-dependent traversal risk. Similarly, terrain-aware models could support real-time adaptation of rover control parameters, such as leg rotation frequency for RHex-type rovers, to optimize speed, stability, or energy and time efficiency across varying levels of regolith compaction. This capability represents a significant step beyond static, pre-mission planning, towards responsive, terrain-aware exploration in uncertain environments.


Finally, our work lays the foundation for long-term, closed-loop exploration, in which legged scouts progressively expand terrain mechanics maps while rovers execute and re-plan trajectories based on updated risk estimates. 
Such adaptation is critical for operating in environments with spatially variable or evolving terrain conditions. In this study, the legged scout followed a predetermined path without explicitly accounting for its own traversal risk. Incorporating safe exploration strategies~\cite{jiang2025safe}, and extending the framework to multi-scout, multi-rover teams would enable dynamic task allocation, more efficient coverage of large exploration areas, increased robustness to individual platform failures, and further enhancement of scientific return through coordinated, terrain-aware decision-making.

\section{Materials and Methods}

\subsection{Hardware and system architecture}\label{sec:method-system}

We deploy a heterogeneous team of robotic platforms with complementary capabilities to evaluate our mission framework:
\textbf{(1)~ Legged scout} (Ghost Robotics Spirit 40, Fig. \ref{fig:overview}-\fbox{1}), a 12-kg quadruped with quasi-direct-drive actuators (1:6 on hip abduction and extension, 1:12 on the knee), representing the highest mobility and proprioceptive sensing capability but with limited payload capacity; 
\textbf{(2)~RHex rover}~\cite{saranli2001rhex} (Fig. \ref{fig:overview}-\fbox{2}), a 8.2 kg hexapedal rover with C-shaped 17.5 cm diameter legs and body dimensions of 54 × 39 x 12 cm
, providing an intermediate balance between mobility and payload capacity; and
\textbf{(3)~Wheeled rover}~\cite{saranli2001rhex} (Fig. \ref{fig:overview}-\fbox{3}), a fixed-axle, four-wheel vehicle measuring 100 × 60 cm, representing a scaled-down analogue of planetary rover designs~\cite{panyam2025trusses} with the highest payload capacity among the platforms.

These platforms were used to validate terrain-informed navigation risk models and assess the effectiveness of planned safe paths.
For Analogue Mission 1, the legged scout was deployed in the NASA Ames lunar regolith testbed to validate the accuracy of the terrain-mapping approach, while the RHex rover was tested in Chrono simulation and laboratory experiment to illustrate the efficacy of the risk-aware planner using the scout-generated regolith map. For Analogue Mission 2, both the legged scout and the wheeled rover were deployed at the White Sands dune field to demonstrate the complete workflow, from terrain mapping to risk-aware rover navigation, in a realistic planetary-analogue environment.

Each robot handles sensing and control onboard using its embedded computer, either a proprietary board paired with an NVIDIA Jetson TX2 (for legged scout) or a Raspberry Pi 4B (for wheeled rover)--and broadcasts messages over ROS 2.
A central mission PC aggregates real-time terrain information and robot state into a Foxglove visual interface~\cite{Foxglove} for operators to facilitate operations (such as monitoring or for setting waypoints).
The system also broadcasts mapping data through ROS 2 network, enabling integration with planning and autonomy modules. 

\subsection{Robotic gaits to enable sensing during locomotion}\label{sec:method-sensing gait}

We employed two different gaits for regolith mapping in this study: a sensing-oriented gait, \textit{Crawl N' Sense}~\cite{fulcher2025effectgaitdesignproprioceptive}, and a locomotion-oriented gait, \textit{Trot-walk}.  

The \textit{Crawl N' Sense} gait~\cite{fulcher2025effectgaitdesignproprioceptive}  was specifically optimized for accurate terrain sensing on deformable substrates. Unlike traditional locomotion-oriented gaits such as trotting or walking, \textit{Crawl N' Sense} maintains at least three legs in contact with the ground to support body weight, enabling a dedicated probing leg to perform controlled interactions with the terrain. This configuration not only enhances the sensing accuracy to estimate the ground reaction forces (GRF) from proprioceptive signals (e.g., current, position) from each motor, but also the robot’s ability to execute different interaction protocols to estimate terrain properties. During experiments at the NASA Ames Research Center, we deployed the \textit{Crawl N' Sense} gait to validate sensing accuracy, where the probing leg was programmed to perform vertical penetrations at a constant velocity of 8 cm/s to accurately measure terrain resistance. 

The \textit{Trot-walk} gait was a baseline locomotion gait that optimizes mobility and speed on natural terrains. For the field deployment at White Sands National Park, we adopted this gait for sensing during locomotion, enabling the robot to scout a much larger area more efficiently. While this approach slightly reduced sensing accuracy due to faster movement, it still provided an acceptable sensing accuracy that has been validated through lab experiments~\cite{fulcher2025effectgaitdesignproprioceptive}. Through our field deployment at White Sands, we show that this gait enabled risk-aware navigation of the wheeled rover to safely reach desired science targets. 

In addition to the two sensing during locomotion gaits, we also conducted independent ground-truth measurements using a standalone robotic-leg penetrometer~\cite{qian2019rapid,ruck2024downslope}. The Traveler leg (Fig. ~\ref{fig:overview}G)  executed controlled penetration experiments to characterize regolith mechanics. The penetration velocity was kept at 1 cm/s to avoid inertial effect~\cite{qian2013walking}, and the penetration depth was kept at 4 cm to avoid boundary effect. These measurements produced high-resolution force–depth profiles that were used to validate the terrain strength estimation.

\subsection{Determining Terrain Properties from Proprioceptively Sensed Forces}\label{sec:method-penetration resistance} 

\begin{figure*}[hb!]
    \centering  
    \includegraphics[width=1\linewidth]{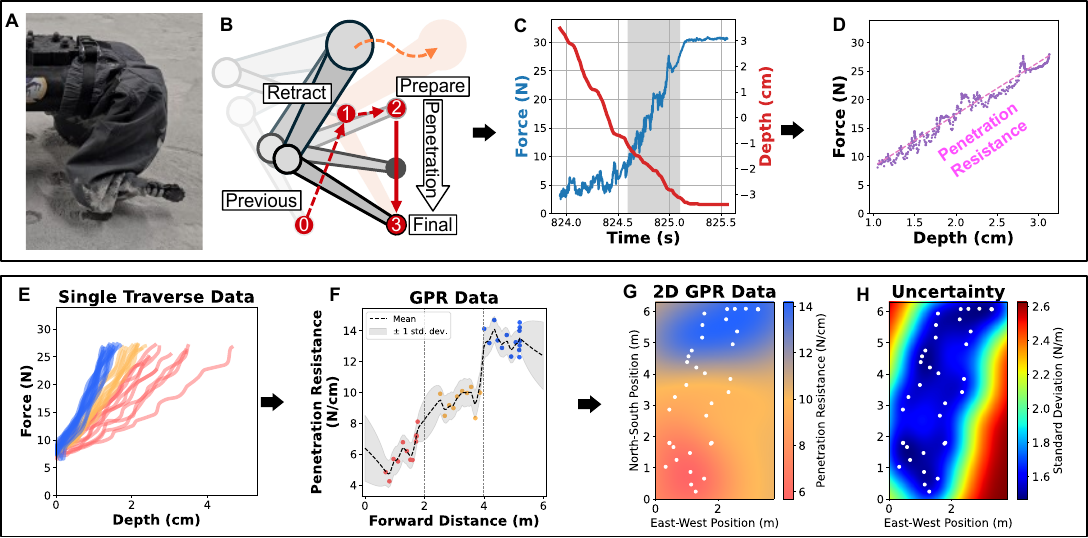}
    \caption{Estimating terrain resistance from legged robot interactions using Gaussian Process regression.
(A, B) A legged scout collects penetration force and toe position data during each stance phase. From the force-depth profile, terrain resistance is computed. (C) Force and position traces are used to estimate ground contact and penetration depth. (D) Penetration resistance is extracted from the slope of the force-depth curve. Bottom: Gaussian Process regression (kriging) is applied for continuous interpolation of terrain strength. Ground truth (E) shows discrete terrain classes. (F) 1D kriging reconstructs resistance along a path with uncertainty bounds. (G) 2D kriging produces a continuous regolith strength map across the testbed, with uncertainty characterization (H).} 
    \label{fig:method1}  
\end{figure*}

To infer intrinsic terrain properties during locomotion, we use proprioceptively measured ground reaction forces (GRFs) and leg penetration depth during each gait cycle’s penetration phase \cite{fulcher2025effectgaitdesignproprioceptive} (Fig. \ref{fig:method1}B, 2–3). Penetration-phase data are extracted from the normal force profile as the foot intrudes into the substrate. To obtain a consistent reference for depth estimation, we estimate a local ground plane using the positions of non-penetrating toes following \cite{fulcher2025effectgaitdesignproprioceptive}, since body motion and joint configuration vary across steps and raw foot position alone does not reflect penetration relative to the true surface height. The penetration depth, z, is then measured normal to this plane. The force and depth data collected during each penetration phase are fit to a granular media model \cite{qian2015principles,li2013terradynamics}. Under this model, the normal penetration resistive force, $f_z$, increases linearly with penetration depth, $z$, scaled by the terrain penetration resistance per unit area, $\resistance$, and the projected surface area of the intruder, $A$.

Accordingly, the force on the robotic leg is modeled as $\hat{f}(z) = A \resistance z$, where $A$ is the projected area of the contact patch on the ground plane. We estimate the penetration resistance per unit area, \resistance, by minimizing the squared error between the modeled and measured forces, $f_z$ (a nominal example is shown in Fig. \ref{fig:method1}D):
\[ \hat \resistance = \operatorname*{arg\,min}_{\resistance} \left( \hat{f}_z - f_z \right)^2 = \operatorname*{arg\,min}_{\resistance} \left( A \resistance z - f_z \right)^2. \]

The resulting penetration resistance reflects an intrinsic terrain property governed by the granular medium’s compaction, composition, and particle size. Additionally, the coefficient of determination, $R^2$, of each linear fit is used as a heuristic for model validity: values near 1 indicate good agreement with the assumed deformable substrate model, while lower values indicate non-linear responses, such as crusted or highly cohesive terrain.



\subsection{Regolith strength map reconstruction}\label{sec:method-GP}
At each step, the robot collects a discrete measurement of the estimated terrain penetration resistance (Fig. \ref{fig:method1}E,F), and the regression error heuristic.
The position of each step in the world frame is then estimated using motion capture (or GPS when outside) and the robot's kinematics.

To predict and map the terrain properties over a continuous 2D area, we use Gaussian Process Regression~\cite{seeger2004gaussian} (GPR), also known as kriging, thereby modeling the underlying terrain property distribution (Fig. \ref{fig:method1}G, H).
By updating the predicted distributions for terrain property map and fit error heuristic map with each step, we provide real-time prediction and quantifiable uncertainty for the terrain properties and the ``trustworthiness" across the landscape for planning traversal and science objectives.
To use GPR, we define the kernel as 
\begin{gather}
    k(x_i,x_j) = k_{C}(x_i,x_j) \times k_{RBF}(x_i,x_j) + k_{WN}(x_i,x_j)    \text{,  with}\\
    k_{RBF}(x_i,x_j) = \exp\left(-\dfrac{d(x_i,x_j)}{2 l^2}\right),\quad
    k_{WN}(x_i,x_j) = \left\{\begin{smallmatrix} n, & x_i=x_j\\ 0, & else \end{smallmatrix}\right., \quad
    k_{C}(x_i,x_j) = C,
\end{gather}%
which is a scaled radial basis function with white noise
where $l, n,$ and $C$ are hyperparameters set based on the environmental priors or optimized to minimize error; descriptions and values in Appendix, Table \ref{tab:gpr_hyperparameters}.


\subsection{Risk Prediction}\label{sec:method-risk}
Measuring the terrain properties directly allow us to better estimate locomotory or operations risk for our rovers.
Risk estimation models were developed for both the RHex-style rovers and our wheeled rovers. 
These models map terrain penetration resistance into mobility outcomes (predicted velocity or wheel slip respectively) and provide prediction of immobilization risk parameterized by platform configuration and payloads.

\subsubsection{RHex rover}\label{sec:risk-RHex}
\begin{figure*}[hb!]
    \centering  
    \includegraphics[width=1\linewidth]{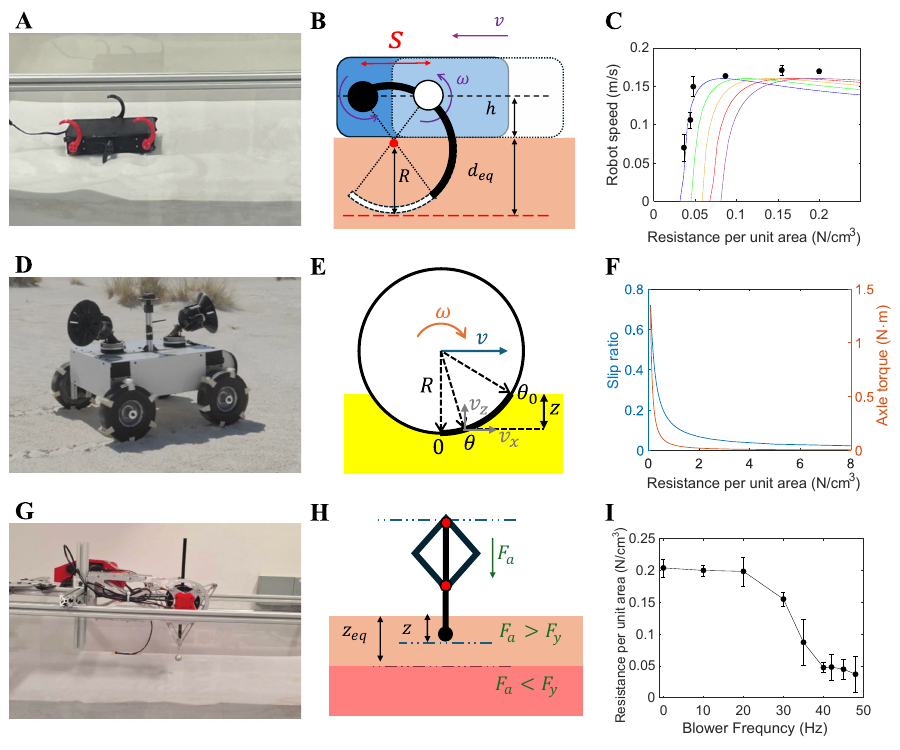}  
    \caption{ 
    Analytical terradynamics models for predicting rover speed. (A) Experimental setup for traversal risk validation using a fluidized granular testbed and a small-scale RHex platform. (B) Terradynamics-based traversal risk model for RHex-class robots~\cite{li2009sensitive,qian2015principles}. (C) Comparison of model-predicted robot speed and experimental measurements. Black markers represent experimentally-measured robot speed for a 0.4 kg rover. Blue, green, orange, red and purple curves represent the model-predicted robot speed for masses of 0.3 kg, 0.4 kg, 0.5 kg, 0.6 kg and 0.7 kg, respectively. (D) Wheeled rover experiments at White Sands analog site. (E) Terradynamics-based traversal risk model for wheeled rovers~\cite{agarwal2019modeling}.  (F) Model-predicted rover slip ratio and axle torque, as a function of terrain penetration resistance per unit area. (G) Traveler to measure the penetration resistance from the fluidized granular testbed. (H) Schematic illustrating the sand solidification depth as a result of comparison between yield force and applied force. (I) Measured terrain penetration resistance per unit area as a function of fluidized bed blower frequency.
} 
    \label{fig:methods2}  
\end{figure*}

Traversal risk for the RHex rover was estimated using a rotary walking model that we previously developed for predicting locomotion performance on granular media~\cite{li2009sensitive,qian2015principles}.
According to this model, upon touchdown, robot legs intrude into the terrain while the sand yields and deforms. As leg intrusion depth increases, the sand resistive force in the vertical direction balances the robot's body weight, at which point the sand locally solidifies and stops yielding further (Fig. \ref{fig:methods2}H, $z_{eq}$). Upon solidification of the sand, The leg then transitions to a rotary walking mode, rotating about its center while propelling the rover forward (Fig. \ref{fig:methods2}B). 

During this sand solidification phase, the penetration depth, $z$, can be expressed as a function of the terrain penetration resistance per unit area, $\alpha_z$, rover parameters (\textit{e.g.,} robot mass $m$) and rover action (\textit{e.g.,} leg angular speed, $\omega$).
\[ z=\frac{m\left[g+\frac{R\omega}{\Delta t}\right]}{n\alpha_zA}=\frac{m\left[g+\frac{R\omega}{\Delta t}\right]}{n\alpha_zRW}, \]
where $R$ and $W$ are the radius and width of the robot C-shaped legs, $\Delta t$ is the characteristic elastic response time of the sand-leg interaction, $g$ is the acceleration due to gravity, and $A=RW$ is the projected surface area of the leg. 
During the rotary walking, 
the hip joint follows a circular trajectory, and the resulting robot fore-aft velocity on sand, $v$, can be computed as a function of the penetration depth (or equivalently, the terrain penetration resistance, Fig. \ref{fig:methods2}C):
\begin{equation}
v=\frac{2R\omega}{\pi}\sqrt{1-\left(\frac{z}{R}+\frac{h}{R}-1\right)^2},
\label{eqn4}
\end{equation}
where $h$ is the robot motor hip height relative to the sand surface (Fig. \ref{fig:methods2}B). 
Based on the predicted robot velocity, $v$,  the slip ratio for a constant leg frequency can be computed as $s =\frac{\frac{2R\omega}{\pi} - v}{v}$. 
In this study, terrain region are considered as high risk for a RHex rover if the slip ratio exceeds 95\%.



\subsubsection{Wheeled rover}\label{sec:risk-wheeled}

Traversal risk for the wheeled rover was estimated using a Resistive Force Theory (RFT) based model\cite{agarwal2019modeling}\cite{agarwal2021surprising}, which predicts the required axle torque for forward motion, and resulting wheel slip ratio, for given terrain strength.

A key advantage of RFT is its ability to handle wheels of arbitrary size and shape, enabling generalization across rover platforms. In our framework, the penetration resistance measured from the scout provides a terrain strength scaling factor, $\zeta \propto \alpha_z$, that modulates the magnitude of granular resistive stresses on flat plate with varying attack and intrusion angles. 
Given rover parameters including wheel diameter and width, RFT predicts wheel forces by linearly superposing local resistive stresses over infinitesimal surface elements of the wheel-terrain contact region.

Using the generic RFT stress distribution\cite{li2013terradynamics}, scaled by the scout-measured scaling factor, $\zeta$, we compute local resistive stresses for arbitrary \emph{attack angle} $\beta$ and \emph{intrusion angle} $\gamma$.
As illustrated in Fig.~\ref{fig:methods2}E, the sinkage depth is represented by the immersion angle $\theta_0$, and velocity components at a local patch with polar angle $\theta$ can be expressed as $v_x(\theta)=v-\omega R\cos\theta$ and $v_z(\theta)=-\omega R\sin\theta$, where $v$ is the wheel linear speed and $\omega$ is the angular speed.
The velocity direction of the local contact patch is determined by the slip ratio, $s = 1 - \frac{v}{\omega R}$.
Thus, the local orientation kinematics can be expressed as
\begin{equation}
\begin{cases}
\beta(\theta, s) = -\theta \\
\gamma(\theta, s) = \operatorname{atan2}(-\sin\theta, 1-s-\cos\theta)
\end{cases}
\end{equation}
Given the orientation parameters $\beta$ and $\gamma$, the stress per unit depth $\alpha_{z,x}$ can be determined for each infinitestimal rim element, the extrusion-induced forces and torque during penetration are then computed using RFT by integrating the stresses over the leading wheel rim surface, $0\le \theta \le \theta_0$:
\begin{equation}
\begin{cases}
F_z(\theta_0, s)=\displaystyle\int_0^{\theta_0}\sigma_z\,\mathrm{d}A\\
F_x(\theta_0, s)=\displaystyle\int_0^{\theta_0}\sigma_x\,\mathrm{d}A\\
\tau(\theta_0, s)=\displaystyle R\int_0^{\theta_0}
\left(\sigma_x\cos\theta+\sigma_z\sin\theta\right)\mathrm{d}A
\end{cases}
\label{eqn:force-torque}
\end{equation}
where the horizontal and vertical components are given by
\[
\sigma_{z,x}(\beta,\gamma)
= \zeta\,\alpha_{z,x}(\beta,\gamma)\,z(\theta),
\]
the infinitesimal contact area is
$\mathrm{d}A = R\ell\,\mathrm{d}\theta$,
where $\ell$ denotes the wheel width,
and the local penetration depth of a rim element is
$z(\theta) = R\bigl(\cos\theta - \cos\theta_0\bigr)$.


Under steady-state conditions, where force equilibrium is achieved, the resulting slip ratio and immersion angle are solved by enforcing force balance in both the vertical and horizontal directions:
\begin{equation}
F_z(\theta_0, s)=mg
\label{eqn:horizontal-force-balance}
\end{equation}
\begin{equation}
F_x(\theta_0, s)=0
\label{eqn:horizontal-force-balance}
\end{equation}
By introducing the dimensionless effective weight $\widetilde{W} \coloneqq \frac{mg}{\zeta R^2 \ell}$, the governing equations can be expressed as follows:
\begin{equation}
\begin{cases}
\int_0^{\theta_0} \alpha_z(\beta,\gamma)(\cos\theta-\cos\theta_0)\mathrm{d}\theta = \widetilde{W}\\
\int_0^{\theta_0} \alpha_x(\beta,\gamma)(\cos\theta-\cos\theta_0)\mathrm{d}\theta=0
\end{cases}
\label{eqn:force-balance-rearranged}
\end{equation}

Therefore, the resulting slip ratio and immersion angle under steady-state advancement can be solved as a function of $\widetilde{W}$.
Substituting the rover parameters, these quantities can be expressed in terms of terrain strength, and the axle torque, $\tau$, can be determined from Eqn. \ref{eqn:force-torque}.

Fig.~\ref{fig:methods2}F shows the computed slip ratio and axle torque as functions of terrain resistance.
Lower terrain strength leads to larger slip ratios, indicating increased immobilization risk, and requires a larger axle torque to sustain forward motion.
Together, slip ratio and axle torque serve as complementary quantitative metrics to evaluate the traversal risk of a given terrain.


\subsection{Lab Validation of the risk model}

To validate the rotary walking–based risk prediction model for RHex-type rovers, we conducted controlled laboratory locomotion experiments using a scaled-down RHex-style prototype (Fig.~\ref{fig:methods2}A). The prototype has a body length of 20.5~cm, a body width of 10.5~cm, and a mass of 0.4~kg. Each C-shaped leg has a diameter of 8~cm and a width of 1.1~cm. Similar to full-scale RHex platforms, the robot employed an alternating-tripod gait~\cite{saranli2001rhex}. In all experiments, the leg rotation frequency was fixed at 1~Hz.

Two sets of experiments were performed to validate the predicted relationship between robot forward speed, terrain strength, and robot parameters. In the first set, we systematically varied the granular compaction of the terrain while keeping robot mass constant. In the second set, we varied the robot payload mass while operating on terrain with a fixed granular compaction. In both cases, experimentally measured robot speeds were compared against model predictions derived from Eq.~\ref{eqn4}.

Terrain compaction was controlled using a sand fluidized bed~\cite{qian2013automated,jin2019preparation}, which allows precise adjustment of granular volume fraction by regulating airflow through the bed. By varying the blower frequency between 0 and 48~Hz, we achieved a wide range of penetration resistance values. For each airflow setting, the resulting penetration resistance per unit area, $\alpha_z$, was measured using a robotic penetrometer (Fig.~\ref{fig:methods2}G) and averaged over 12 spatially distributed locations on the sand surface. As blower frequency increased, $\alpha_z$ decreased monotonically from $0.204\ \mathrm{N/cm^3}$ to $0.039\ \mathrm{N/cm^3}$, corresponding to progressively looser sand conditions, shown in Fig.~\ref{fig:methods2}I.


Prior to each trial, the sand bed was fully fluidized by increasing airflow beyond the onset of fluidization~\cite{jin2019preparation}, then slowly defluidized to the target blower frequency to ensure a level and repeatable initial surface condition. For each combination of terrain strength and robot mass, three trials were conducted, each lasting 10~s. Robot position was tracked using a motion-capture system, and average forward speed was computed over the steady-state portion of each trial.

Experimental results show that the robot maintained a nearly constant forward speed of approximately $v$ = 0.16 m/s on relatively compacted sand ($\alpha_z >0.1~\rm{N/cm^3}$). As terrain strength decreased, the robot experienced a sharp performance degradation, ultimately failing to make forward progress near the model-predicted critical penetration resistance $\alpha_z^*=0.05~\rm{N/cm^3}$. Overall, the measured robot speeds closely matched theoretical predictions across the tested terrain conditions (Fig.~\ref{fig:methods2}C), indicating that the model accurately captures the onset of locomotion failure. These results confirm that the robot can safely traverse sufficiently compacted sand, while identifying softer substrates that pose a high immobilization risk.

Additional experiments with increased robot mass further support the model predictions. As robot mass increased from 0.3~kg to 0.7~kg, the critical penetration resistance threshold shifted to larger values, indicating that heavier robots require stiffer terrain to maintain mobility. This trend is consistent with the rotary walking model: increased normal load necessitates deeper leg intrusion to balance weight, leading to reduced forward speed and a higher likelihood of failure on soft substrates. 





\subsection{Autonomous mission execution through potential-field-based navigation}

The rover, which carries the scientific payload, is tasked with minimizing traversal risk while visiting regions identified as scientifically valuable by the science team. As the scouting robot explores the environment, it produces two additional maps based on the regolith strength to support this objective: (1) a traversal risk map that predicts the likelihood that a rover will get stuck, and (2) a scientific reward map that represents the likelihood that the scientific measurements can help validate or invalidate a specific hypothesis.

The traversal risk map is generated
by utilizing the locomotion model described in Sec.~\ref{sec:method-risk}.
The scientific reward map is computed using methods from our prior work~\cite{liu2023understanding} and is provided to the science team to support the identification of candidate sampling locations.

To demonstrate the ability of the system to use the traversal risk map to navigate safely to areas of high scientific interest, we implement an artificial potential field planner. Specifically, the traversal risk map was thresholded and converted into a binary obstacle map consisting of polygonal obstacles.
In the potential field planner~\cite{hwang1992potential}, each obstacle then corresponds to a repulsive field 
\begin{equation}
F_{rep}=\frac{k_{rep}}{D(p_{r})^2}\hat{n}(p_{r})
\end{equation}
where $k_{rep}=0.1$ is a scalar gain that was tuned manually for this environment, $D$ is the rover's distance to the obstacle, $\hat{n}$ is a unit vector pointing from the closest point on the obstacle boundary to the rover, and both $D$ and $\hat{n}$ are functions of the rover's current location $p_r$.
In addition, each location of scientific interest was converted into an attractive potential field of the form
\begin{equation}
F_{att}=k_{att}(p_{goal}-p_{rover})
\end{equation}
where $k_{att}=1$ is a manually tuned scalar gain and $p_{goal}$ is the current goal location of interest.
The scientific locations were manually sequenced so that only one goal location was used at a time.
The rover itself was modeled as a point mass with a maximum velocity of 1 m/s and a maximum acceleration of 1 m/s$^2$. In this way, the generated path could include forward, backward, and limited turning consistent with the hardware platform.

\begin{figure*}[hb!]
    \centering  
    \includegraphics[width=1\linewidth]{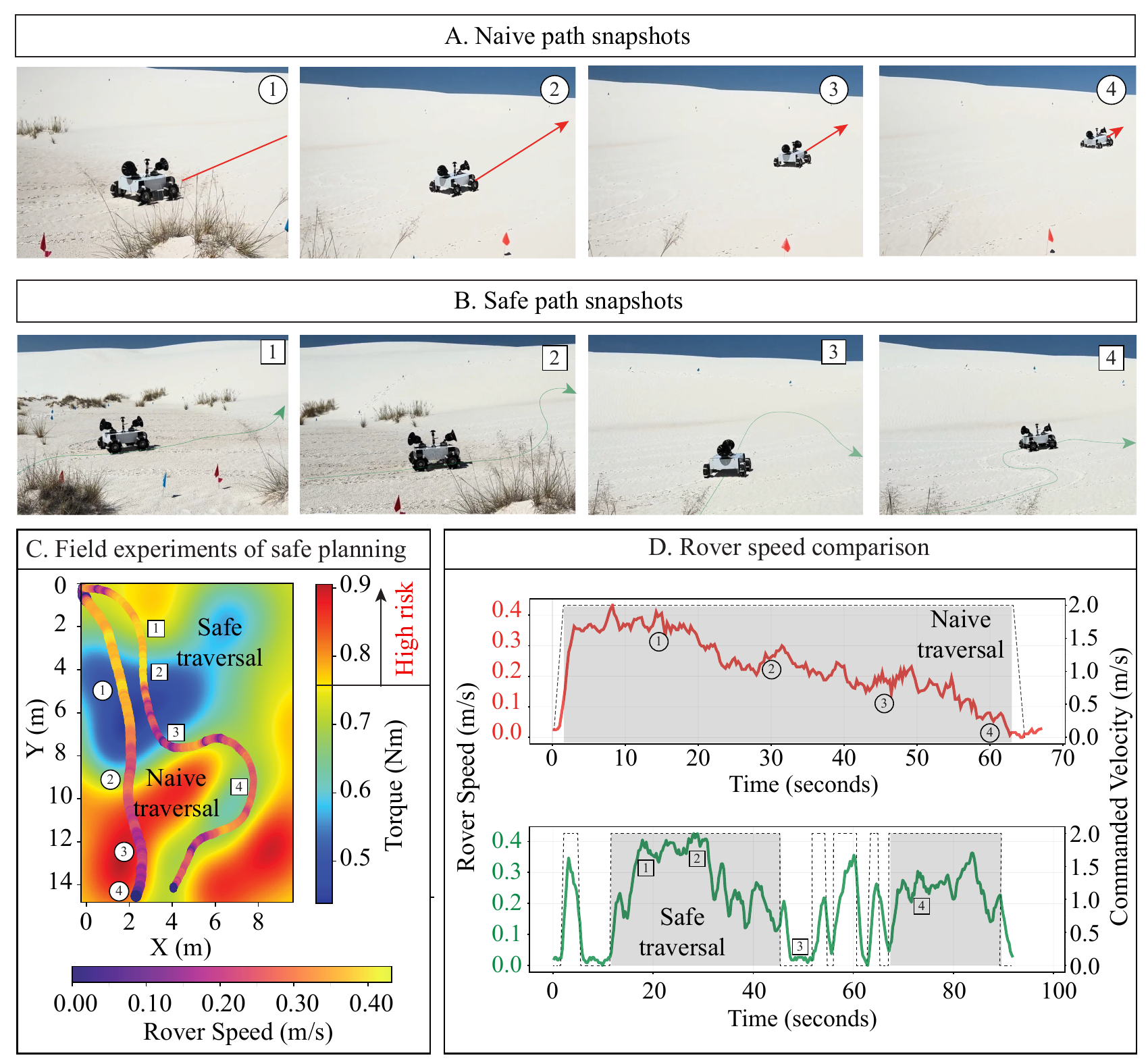}  
    \caption{Comparison of naive versus safe path planning for rover navigation.  
(A) Snapshots of rover traversal along the naive path at (1) $T=15$ s, (2) $T=30$ s, (3) $T=45$ s, and (4) $T=60$ s, where the rover speed drops to near zero upon entering a soft region.  
(B) Snapshots of rover traversal along the safe path at (1) $T=20$ s, (2) $T=30$ s, (3) $T=48$ s, and (4) $T=72$ s, where the rover maintains acceptable forward speed.  
(C) Predicted rover traversal risk (based on required torque) with the naive path (dashed black) and the safe path (green) overlaid. The color of the path represent experimentally-measured rover speed.   
(D) Time series of commanded versus actual rover speed during the two traversals. Labels (1)–(4) in A–D correspond to the same waypoints along the path.  
} 
    \label{fig:planning}  
\end{figure*}

To generate the path, at each point in time, the attractive and repulsive forces were summed and clamped so that the net acceleration of the rover would not exceed the maximum acceleration magnitude. This acceleration was then applied to the rover's current velocity, and the velocity was then clamped to the maximum velocity.
The resulting path, generated using time steps of 0.01~s, is shown in Fig. \ref{fig:planning}C, safe traversal path. 


\clearpage

	

\vspace{-1in}


\clearpage 

%
\bibliography{main} 
\bibliographystyle{sciencemag}

%
%
%
%
%
%


\section*{Acknowledgments}
The authors would like to thank the NASA Ames Research Center and NASA Solar System Exploration Research Institute (SSERVI) for the access to the regolith testbed, and Dr. Allison Okamura for the use of her motion capture system. The authors would also like to thank Raymond Yang and Marissa Teitelbaum for helping with rover development; Natalie Cavallo, Xinyue Guo, Matthew Jiang, Julian Maynes for helping with algorithm implementation; and Swarnadip Saha for helping with preliminary data collection.


\paragraph*{Funding:}
 This project is funded by the NASA Lunar Surface Technology Research (LuSTR) program, Award \# 80NSSC24K0127, the NASA Planetary Science and Technology Through Analog Research (PSTAR) program, Grant 80NSSC22K1313, the National Science Foundation (NSF) CAREER Award \# 2240075, National Science Foundation (NSF) Foundational Research in Robotics (FRR) program, Award \# 2529696, and the NASA Mars Exploration Program (MEP) Technology Development Funding.

\paragraph*{Data and materials availability:} 
All data needed to evaluate the conclusions in the paper are present in the paper and/or the Supplementary Materials. The datasets and analysis code supporting the findings of this study are available at GitHub: \\
\url{https://github.com/QianLabUSC/Scout-Rover-Cooperation}



\newpage
\section*{Appendix A: Simulation Parameter Mapping}\label{sec:appendix}

\subsection*{A.1 Stiffness Parameter Matrix}

The spatial parameterization of the NASA Ames LHS-1 simulant testbed is implemented through a 6×4 stiffness parameter matrix, where each cell corresponds to a 1×1 meter region of the physical testbed. The stiffness values (in N/cm³) are derived from ground-truth measurements collected by the Traveler leg penetrometer and represent the average penetration resistance for each grid cell.

\begin{table}[h]
\centering
\caption{\textbf{Stiffness parameter matrix for NASA Ames LHS-1 simulant testbed.} Each cell represents the average stiffness value (N/cm$^3$) for a 1$\times$1 meter region, with coordinates (x, y) corresponding to the physical testbed layout.}
\label{tab:stiffness_matrix}

\begin{tabular}{|c|c|c|c|c|}
\hline
\textbf{Y/X} & \textbf{0-1m} & \textbf{1-2m} & \textbf{2-3m} & \textbf{3-4m} \\
\hline
\textbf{0-1m} & 4.17 & 1.16 & 0.31 & 0.74 \\
\hline
\textbf{1-2m} & 4.93 & 1.04 & 0.30 & 0.28 \\
\hline
\textbf{2-3m} & 4.46 & 5.75 & 1.41 & 1.65 \\
\hline
\textbf{3-4m} & 1.23 & 4.04 & 3.93 & 1.71 \\
\hline
\textbf{4-5m} & 0.35 & 0.35 & 5.98 & 4.87 \\
\hline
\textbf{5-6m} & 1.03 & 0.56 & 8.38 & 4.16 \\
\hline
\end{tabular}
\end{table}

\subsection*{A.2 Parameter Mapping Methodology}

The stiffness values in Table \ref{tab:stiffness_matrix} are mapped to corresponding elastic and damping parameters in the Chrono physics engine simulation using the SCM (Soil Contact Model) terrain representation. The mapping function \texttt{getMatrixValues(x, y)} interpolates the stiffness values based on the robot's position coordinates, where:

\begin{itemize}
\item \textbf{X-coordinate range}: 0-4 meters (4 columns)
\item \textbf{Y-coordinate range}: 0-6 meters (6 rows)  
\item \textbf{Stiffness range}: 0.28 to 8.38 N/cm$^3$
\item \textbf{Cell size}: 1$\times$1 meter
\end{itemize}

The stiffness values are scaled and mapped to SCM terrain parameters as follows:
\begin{itemize}
\item \textbf{Bekker\_Kphi}: $matrix\_value \times 10^6$ (scaled stiffness)
\item \textbf{Bekker\_Kc}: -100$\times$10$^3$ (cohesion modulus)
\item \textbf{Bekker\_n}: 1.0 (sinkage exponent)
\item \textbf{Mohr\_cohesion}: 1.3$\times$10$^3$ (cohesion)
\item \textbf{Mohr\_friction}: 31.1° (friction angle)
\item \textbf{Janosi\_shear}: 1.2$\times$10$^{-2}$ (shear deformation modulus)
\item \textbf{elastic\_K}: $8 \times stiffness$ (elastic stiffness)
\item \textbf{damping\_R}: 5$\times$10$^4$ (damping coefficient)
\end{itemize}

The stiffness values reflect the three distinct terrain preparation protocols:
\begin{itemize}
\item \textbf{High stiffness (4.0-8.4 N/cm$^3$)}: Tamped regions (red in simulation)
\item \textbf{Medium stiffness (1.0-4.0 N/cm$^3$)}: Raked regions (yellow in simulation)  
\item \textbf{Low stiffness (0.3-1.0 N/cm$^3$)}: Sifted regions (blue in simulation)
\end{itemize}

This spatial parameterization enables the simulation to accurately reproduce the physical testbed conditions, providing a validated environment for testing rover traversal risk assessment algorithms.



\begin{table} 
	\centering
	\caption{\textbf{Gaussian Process Regression (GPR) hyperparameters.}
		Physical descriptions, if applicable, and values of the hyperparameter used a}
	\label{tab:gpr_hyperparameters} 
	
	\begin{tabular}{lcccl} 
		\\
		\hline
		Hyperparameter & Symbol & Value & Bounds & Description\\
		\hline
		Length Scale & $l$ & 0.5 &\\
		Noise Floor& $n$ & 0.2&\\
		Constant & $c$ &  5 &\\
		\hline
	\end{tabular}
\end{table}


\end{document}